\documentclass{article}
\usepackage{spconf}
\usepackage{amsmath,graphicx,hyperref}
\usepackage{multirow}
\usepackage{booktabs}
\usepackage{float}
\usepackage{enumitem}
\usepackage{makecell}

\usepackage{amsfonts}
\usepackage{tabularx}
\usepackage{subcaption}
\usepackage{svg}
\usepackage{cite, hyperref}

\title{Dual-level Progressive Hardness-Aware Reweighting for Cross-View Geo-Localization}

\name{
  \shortstack{
    Guozheng Zheng$^{1}$, 
    Jian Guan$^{1,*}$\thanks{*Corresponding authors. \\This work was supported by the project under Grant No. D040303.},
    Mingjie Xie$^{2}$, 
    Xuanjia Zhao$^{1}$, 
    Congyi Fan$^{1}$, \\
    Shiheng Zhang$^{1}$,
    Pengming Feng$^{3,*}$
  }
}

\address{
    $^1$ Group of Intelligent Signal Processing, Harbin Engineering University, Harbin, China \\
    $^2$ School of Astronautics, Beihang University, Beijing, China \\
    $^3$ State Key Laboratory of Space Information System and Integrated Application, Beijing, China
}

\begin{document}

\maketitle

\begin{abstract}
Cross-view geo-localization (CVGL) between drone and satellite imagery remains challenging due to severe viewpoint gaps and the presence of hard negatives, which are visually similar but geographically mismatched samples. Existing mining or reweighting strategies often use static weighting, which is sensitive to distribution shifts and prone to overemphasizing difficult samples too early, leading to noisy gradients and unstable convergence. In this paper, we present a Dual-level Progressive Hardness-aware Reweighting (DPHR) strategy. At the sample level, a Ratio-based Difficulty-Aware (RDA) module evaluates relative difficulty and assigns fine-grained weights to negatives. At the batch level, a Progressive Adaptive Loss Weighting (PALW) mechanism exploits a training-progress signal to attenuate noisy gradients during early optimization and progressively enhance hard-negative mining as training matures. Experiments on the University-1652 and SUES-200 benchmarks demonstrate the effectiveness and robustness of the proposed DPHR, achieving consistent improvements over state-of-the-art methods.

\end{abstract}


\section{Introduction}
\label{sec:intro}

Cross-view geo-localization (CVGL) between drone and satellite imagery aims to retrieve the geographically corresponding image from a gallery in another view given a query image \cite{Lin2013cross}. It is a fundamental task for applications such as aerial inspection, autonomous navigation, and urban-scale delivery~\cite{Zhu2018vision, Humenberger2022investigating, yu2019building}. Despite its importance, the task remains highly challenging due to severe viewpoint discrepancies, scale variations, and appearance differences caused by altitude and imaging conditions \cite{Zheng2020University1652}.

\begin{figure}[t]
  \centering
  \includegraphics[width= .9\linewidth]{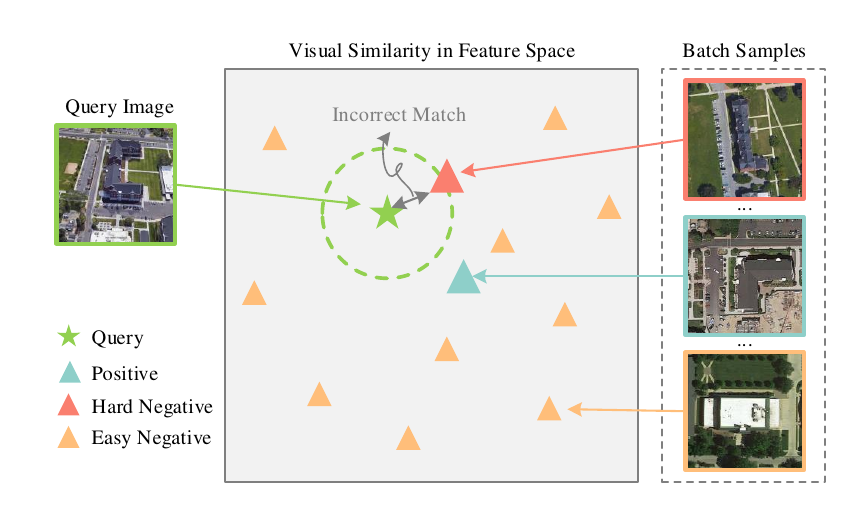}
  \caption{
  Illustration of visual similarity in feature space for CVGL. Given a query image, the challenge arises when the hard negative, due to its structural and color similarities, becomes closer to the query in feature space than the true positive, misguiding the model into making incorrect matches.
  }
  \label{fig:Hard-vis}
  \vspace{-.4cm}
\end{figure}

The aforementioned challenges give rise to hard negatives, \textit{i.e.}, samples that are geographically mismatched yet visually similar to the query, which pose a major obstacle for CVGL.
As illustrated in Fig.~\ref{fig:Hard-vis}, such negatives may even appear closer to the query than the true positive in the feature space. This not only misguides the model into incorrect matches but also causes them to dominate gradient updates, destabilizing training and hindering convergence.

Existing work has sought to mitigate this issue through either stronger representation learning or targeted hard-negative handling~\cite{triplet, soft-margin-triplet, infonce}. For instance, sampling-based methods~\cite{HADGEO, Sample4Geo, Effective-Negative-Sampling} expose the model to visually confusing pairs to strengthen discrimination. Loss-based reweighting approaches, exemplified by HER~\cite{Cai2019GroundToAerial}, assign larger weights to difficult triplets using gap-based functions with stability controls. While these strategies demonstrate effectiveness, they remain limited in three key respects. First, static difficulty definitions are sensitive to distribution shifts across scenes, leading to inconsistent weighting of equally difficult samples. Second, clipping strategies, \textit{e.g.}, ~\cite{Cai2019GroundToAerial}, collapse extremely hard cases to the same threshold, erasing fine-grained distinctions among the most confusing negatives. Third, time-invariant weighting prematurely emphasizes hard negatives in early training, when representations are still crude, thereby amplifying noise and reducing overall retrieval performance.

To address these issues, we propose a dual-level progressive hardness-aware reweighting (DPHR) strategy for CVGL. Specifically, at the sample level, we introduce a ratio-based difficulty-aware (RDA) module that adaptively allocates weights according to the relative hardness of negatives, ensuring consistent emphasis even under varying data distributions. At the batch level, we design a progressive adaptive loss weighting (PALW) mechanism that leverages a training-progress signal derived from recent unweighted losses to dynamically regulate the influence of difficult samples. This progressive adjustment suppresses noise during the unstable early phase and gradually strengthens hard-negative mining as training stabilizes, achieving a balance between robustness and discriminability. Extensive experiments on two drone–satellite benchmarks, \textit{i.e.}, University-1652~\cite{Zheng2020University1652} and SUES-200~\cite{Zhu2023SUES200}, confirm the effectiveness of our approach. The proposed strategy consistently enhances Recall@1 and Average Precision across different retrieval directions, demonstrating the effectiveness and improved robustness compared with state-of-the-art methods.

\section{Proposed Method}

\begin{figure}[t]
    \centering
    \includegraphics[width=1\linewidth]{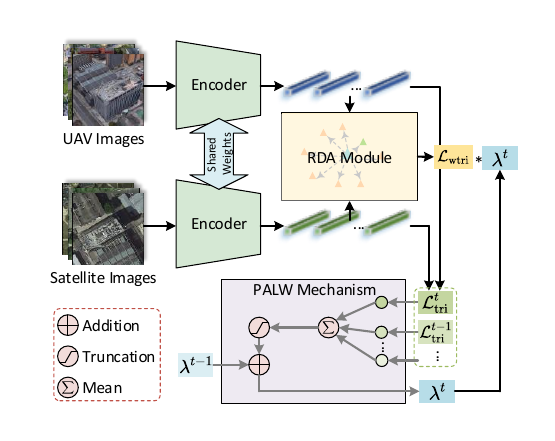}
    \caption{
    The overall framework of the proposed DPHR strategy for CVGL, which consists of two key components, \textit{i.e.}, ratio-aware difficulty-aware (RDA) module and progressive adaptive loss weighting (PALW) mechanism. Here, RDA module assigns sample-level weights based on relative hardness, while the PALW mechanism adaptively regulates the overall loss contribution according to training progress. 
    }
    \label{fig:backbone}
\end{figure}

To tackle the challenges posed by hard negatives in CVGL, we introduce a dual-level progressive hardness-aware reweighting (DPHR) strategy. The overall framework is illustrated in Fig.~\ref{fig:backbone}, which comprises two complementary components, \textit{i.e.}, ratio-based difficulty-aware (RDA) module and progressive adaptive loss weighting (PALW) mechanism.

\subsection{Preliminaries}

We follow the standard CVGL setting, where the goal is to retrieve the geographically corresponding image in another view for a given query image. A weight-sharing dual-branch encoder $\mathcal{F}(\cdot)$ is adopted to extract robust image embeddings~\cite{Shen2023MCCG, ConvNeXt-based, ConvNeXt-shaa}. Specifically, given a batch of cross-view paired images, \textit{i.e.}, $\{x_i^j\}_{i=1}^{B}$, the encoder outputs image embedding $e_i^j=\mathcal{F}(x_i^j)$, where $B$ is the batch size and $j\in\{\text{Drone}, \text{Satellite}\}$ denotes the platform. 

Assuming the embedding of $i$-th sample is selected as the query embedding, we construct a triplet of $(q_i, p_i, n_i)$, where $q_i$ denotes the query embedding, $p_i$ indicates its corresponding positive embedding and $n_{i,k}$ is the $k$-th negative embedding. Thus, the original triplet loss is formulated as follows:
\begin{equation}
  \mathcal{L}_{\text{tri}} = \frac{1}{B(B-1)} \sum_{i=1}^B \sum_{k=1}^{B-1} \ell_{\text{tri}}(i,k), \\
\end{equation}
\begin{equation}
  \ell_{\text{tri}}{(i,k)} = \max \left(0, d(q_i,p_i)-d(q_i,n_{i,k})+m \right), 
\end{equation}
where $d(\cdot)$ represents the squared Euclidean distance between two inputted embeddings, and $m$ is the max-margin to enforce a minimum separation between $d(q_i,p_i)$ and $d(q_i,n_{i,k})$.

\subsection{Ratio-Based Difficulty-Aware Module}
\label{ssec:rhw}
In order to emphasize informative hard negatives while suppress easy ones, we introduce the RDA module to provide a normalized hardness score $h_{i,k}$ for each negative sample:
\begin{equation}
\label{eq:das}
h_{i,k}=\frac{d{(p_i,q_i)}}{\,d{(p_i,q_i)}+d{(q_i,n_{i,k})}}\in[0,1],
\end{equation}
where the larger $h_{i,k}$ indicates that the negative is closer to the query relative to the positive, which is therefore more difficult. This ratio is scale-invariant under global distance rescaling, avoiding weight drift caused by varying feature norms.
In addition, to ensure sufficient emphasis on hard negatives, we map $h_{i,k}$ linearly to a weight interval $[w_{\min},w_{\max}]$:
\begin{equation}
\begin{aligned}
 \omega_{i,k} & = \operatorname{scale}(w_{\min},w_{\max},h_{i,k})\\
  & =w_{\min}+(w_{\max}-w_{\min})\,h_{i,k},
\end{aligned}
\end{equation}
where $\operatorname{scale}(\cdot)$ is a linear scaling function. After that, the hardness-weighted triplet loss can be formulated as:
\begin{equation}
\mathcal{L}_{\mathrm{wtri}}=\frac{1}{B(B-1)}\sum_{i=1}^B\sum_{k=1}^{B-1} \omega_{i,k}\,\ell_{\text{tri}}{(i,k)},
\end{equation}
where difficult negative samples receive a larger gradient contribution while easy ones are down-weighted.

\subsection{Progressive Adaptive Loss Weighting Mechanism}
\label{ssec:talw}
Although hard negatives are crucial, overweighting them at early training stages can introduce noisy gradients due to unstable embeddings. To address this, our PALW mechanism is proposed to adaptively scale the hardness-weighted loss $\mathcal{L}_{\mathrm{wtri}}$ based on training progress.

Specifically, for the $t$-th iteration, we first compute a progress signal $\alpha_t$ as the moving average of the unweighted triplet loss over the most recent $R_{t}$ iterations:
\begin{equation}
\alpha_t=\frac{1}{R_t}\sum_{r=0}^{R_t-1}\mathcal{L}_{\text{tri}}^{t-r},
\end{equation}
where $R_t=\min(R,t+1)$ is the window size to select an appropriate number of recent losses and $R$ is the preset maximum window size. This signal is then normalized as follows:
\begin{equation}
\hat\alpha_t=\operatorname{trunc}\!\left(\frac{\alpha_t-\sigma_{\min}}{\sigma_{\max}-\sigma_{\min}},0,1\right),
\end{equation}
where $\operatorname{trunc}(\cdot)$ indicates the truncation operation to ensure that the normalized process signal $\hat\alpha_t$ falls within the range of $[0,1]$. An instantaneous scaling coefficient is defined as:
\begin{equation}
\lambda_{\mathrm{inst}}=\operatorname{scale}(\delta_{\min},\delta_{\max},(1-\hat\alpha_t)^\gamma),
\end{equation}
where $\gamma$ controls the transition rate, empirically set to 1.5. At early stages, $\hat\alpha_t$ is relatively large, so $\lambda_{\mathrm{inst}}$ is close to $\delta_{\min}$, suppressing noisy hard negatives. As training stabilizes, $\hat\alpha_t$ decreases and $\lambda_{\mathrm{inst}}$ approaches $\delta_{\max}$, amplifying hard-negative contributions.

In addition, to smooth short-term fluctuations and capture long-term trends, we apply an exponential moving average (EMA) operation as follows:
\begin{equation}
\lambda^{t}=\beta\,\lambda^{t-1}+(1-\beta)\lambda_{\mathrm{inst}},
\end{equation}
where $\beta$ is a smoothing factor with a default value of 0.9. Thus, the final objective can be expressed as:
\begin{equation}
\mathcal{L}_{\text{DPHR}}=\mathcal{L}_{\text{tri}}+\lambda^{t}\,\mathcal{L}_{\mathrm{wtri}},
\end{equation}
which achieves progressive adaptation from robustness in the early stages to enhanced discriminability later.

\section{Experiments}
\label{sec:experients}
\subsection{Experimental Setup}
\noindent\textbf{Datasets:} In order to evaluate the effectiveness of our proposed strategy, we conduct experiments on two widely used drone-satellite CVGL benchmarks, \textit{i.e.}, University-1652~\cite{Zheng2020University1652} and SUES-200~\cite{Zhu2023SUES200}. Specifically, the University-1652 dataset consists of images from three platforms, \textit{i.e.}, drone, satellite and street-view, covering 1,652 buildings across 72 universities. Following the standard protocol \cite{Shen2023MCCG, Sample4Geo, LPN, University-method02}, 701 buildings from 33 universities are used for training, while 951 buildings from the remaining 39 universities are reserved for testing. The SUES-200 dataset contains drone and satellite imagery from 200 locations. The drone images are captured at four altitudes (\textit{i.e.}, 150m, 200m, 250m, and 300m), with 50 drone images per altitude. Each location is paired with one corresponding satellite image, enabling evaluation under diverse viewpoint and scale variations.

\noindent\textbf{Evaluation Metrics:} For fair comparison, we follow prior works~\cite{Shen2023MCCG, Sample4Geo, LPN} and adopt two standard retrieval metrics, \textit{i.e.}, Recall@1 (denoted as R@1) and Average Precision (AP). Here, R@1 measures the percentage of queries whose top-1 retrieved result is the correct match, directly reflecting the accuracy of retrieval. AP corresponds to the area under the Precision–Recall curve, capturing the balance between precision and recall across different thresholds, thus offering a more comprehensive evaluation of CVGL performance. In our experiments, we evaluate performance under two retrieval directions, namely \emph{Drone$\rightarrow$Satellite} and \emph{Satellite$\rightarrow$Drone}.

\noindent\textbf{Implementation Details:}
To thoroughly validate the effectiveness and robustness of the proposed strategy, we integrate it into three representative CVGL frameworks, \textit{i.e.}, LPN~\cite{LPN}, MCCG~\cite{Shen2023MCCG}, and Sample4Geo~\cite{Sample4Geo}. All training configurations and backbones strictly follow the original implementations to ensure a fair comparison. 
For our strategy, the triplet margin is set to $m=0.3$ and the training-state normalization adopts $\sigma_{\min}=0.8$ and $\sigma_{\max}=1.5$. As for the linear scaling function, the sample-level difficulty weight $\omega$ is constrained within $[0.5, 2.0]$, and the stabilized batch-level coefficient $\lambda_\mathrm{inst}$ is bounded within $[0.2, 1.0]$.

\subsection{Performance Comparison}
\label{ssec:subhead}

To evaluate the effectiveness of the proposed DPHR strategy in enhancing CVGL, we integrate it with several representative methods, namely LPN-DPHR, MCCG-DPHR, and Sample4Geo-DPHR. Tables~\ref{tab:u1652_single} and \ref{tab:sues200_single} report results on the University-1652 and SUES-200 datasets, respectively.

\begin{table}[!b]
\centering
\caption{Performance comparison on University-1652.}
\label{tab:u1652_single}
\renewcommand{\arraystretch}{1.0}
\resizebox{0.9\linewidth}{!}{
\setlength{\tabcolsep}{2pt}
\begin{tabular}{l c c c c}
\toprule
\multirow{2}{*}{\textbf{Method}} &
\multicolumn{2}{c}{\textbf{Drone $\rightarrow$ Satellite}} &
\multicolumn{2}{c}{\textbf{Satellite $\rightarrow$ Drone}} \\
\cmidrule(lr){2-3} \cmidrule(lr){4-5}
 & R@1 & AP & R@1 & AP \\
\midrule
LPN~\cite{LPN} & 74.19 & 77.55 & 85.02 & 73.24 \\
\textbf{LPN-DPHR}& \textbf{74.62} & \textbf{77.87}  
            & \textbf{86.73} & \textbf{74.93} \\
\midrule
MCCG~\cite{Shen2023MCCG} & 88.58 & 90.37 & 93.87 & 88.82 \\
\textbf{MCCG-DPHR} & \textbf{89.29} & \textbf{90.97}  
             & \textbf{95.15} & \textbf{89.41} \\
\midrule
Sample4Geo~\cite{Sample4Geo} & 92.05 & 93.36 & 94.29 & 88.44 \\
\textbf{Sample4Geo-DPHR} & \textbf{92.32} & \textbf{93.62}  
                  & \textbf{94.44} & \textbf{89.27} \\
\bottomrule
\end{tabular}}
\end{table}
\begin{table}[t]
\centering
\caption{Performance comparison on SUES-200.}
\label{tab:sues200_single}
\setlength{\tabcolsep}{2pt}
\resizebox{\linewidth}{!}{
\begin{tabular}{lcccccccc}
\toprule
\multirow{2}{*}{\textbf{Method}} &
\multicolumn{2}{c}{\textbf{150m}} &
\multicolumn{2}{c}{\textbf{200m}} &
\multicolumn{2}{c}{\textbf{250m}} &
\multicolumn{2}{c}{\textbf{300m}} \\
\cmidrule(lr){2-3} \cmidrule(lr){4-5} \cmidrule(lr){6-7} \cmidrule(lr){8-9}
& R@1 & AP & R@1 & AP & R@1 & AP & R@1 & AP \\
\midrule
\multicolumn{9}{c}{\textbf{Drone $\rightarrow$ Satellite}} \\
\midrule
LPN~\cite{LPN} & 52.90 & 59.29 & 63.88 & 69.58 & 73.45 & 78.14 & 85.08 & 87.79 \\
\textbf{LPN-DPHR} & \textbf{58.69 }& \textbf{62.43} & \textbf{75.22} & \textbf{79.80} & \textbf{76.32} & \textbf{80.68} & \textbf{85.72} & \textbf{88.64} \\
MCCG~\cite{Shen2023MCCG} & 78.85 & 82.60 & 89.67 & 91.67 & 94.60 & 95.71 & 96.10 & 96.86 \\
\textbf{MCCG-DPHR} & \textbf{84.05} & \textbf{87.18} & \textbf{91.28} & \textbf{92.98} & \textbf{95.00} & \textbf{95.92} & \textbf{96.95} & \textbf{97.33} \\
Sample4Geo~\cite{Sample4Geo} & 86.23 & 88.55 & 92.45 & 93.79 & 97.02 & 97.56 & 98.25 & 98.66 \\
\textbf{Sample4Geo-DPHR} & \textbf{94.55} & \textbf{95.60} & \textbf{95.43} & \textbf{96.36} & \textbf{98.95} & \textbf{99.14} & \textbf{99.80} & \textbf{99.85} \\
\midrule
\multicolumn{9}{c}{\textbf{Satellite $\rightarrow$ Drone}} \\
\midrule
LPN~\cite{LPN} & 67.50 & 70.71 & 86.25 & 73.84 & 86.25 & 73.68 & 85.05 & 87.77 \\
\textbf{LPN-DPHR} & \textbf{70.25} & \textbf{72.43} & \textbf{91.25} & \textbf{85.28} & \textbf{88.75} & \textbf{79.74} & \textbf{100.00} & \textbf{92.10} \\
MCCG~\cite{Shen2023MCCG} & 92.50 & 83.44 & 97.50 & 92.01 & 96.25 & 95.78 & 97.50 & \textbf{96.98} \\
\textbf{MCCG-DPHR} & \textbf{95.00} & \textbf{87.60} & \textbf{97.50} & \textbf{92.98} & \textbf{97.50} & \textbf{96.72} & \textbf{97.50} & 96.82 \\
Sample4Geo~\cite{Sample4Geo} & 95.00 & 84.47 & 96.25 & 91.56 & 97.50 & 95.25 & 98.75 & 96.69 \\
\textbf{Sample4Geo-DPHR} & \textbf{95.00} & \textbf{90.73} & \textbf{97.50} & \textbf{94.41} & \textbf{98.75} & \textbf{97.70} & \textbf{99.88} & \textbf{99.90} \\
\bottomrule
\end{tabular}
}
\end{table}
Across all retrieval directions and altitudes, DPHR consistently improves performance in terms of R@1 and AP, demonstrating broad applicability and robustness across different CVGL models. The improvements are especially notable under challenging conditions, such as the 150m drone altitude on SUES-200. For example, in the Drone$\rightarrow$Satellite task, applying DPHR to Sample4Geo increases R@1 from 86.23 to 94.55 and AP from 88.55 to 95.60. At low drone altitudes, the larger viewpoint gap and smaller ground footprint lead to missing or occluded discriminative cues, generating more hard negatives. By difficulty-aware reweighting, DPHR amplifies informative contrasts while suppressing early-stage noise, effectively improving retrieval accuracy. In contrast, under high-altitude settings or the Satellite$\rightarrow$Drone direction, improvements are relatively smaller, as the increased similarity between drone and satellite images reduces the relative difficulty of negative samples.

\subsection{Ablation Study}
\begin{table}[t]
\centering
\caption{Results of ablation study using MCCG as baseline on University-1652.}
\setlength{\tabcolsep}{6pt}
\resizebox{\linewidth}{!}{
\begin{tabular}{cc|cc|cc}
\toprule
& & \multicolumn{2}{c|}{\textbf{Drone $\rightarrow$ Satellite}} & \multicolumn{2}{c}{\textbf{Satellite $\rightarrow$ Drone}} \\
RDA & PALW & R@1 & AP & R@1 & AP \\
\midrule
$\times$ & $\times$ & 88.58 & 90.37 & 93.87 & 88.82 \\
\checkmark & $\times$ & 85.47 & 87.73 & 92.58 & 85.87 \\
$\times$ & \checkmark & 89.01 & 90.71 & 94.15 & 89.17 \\
\checkmark & \checkmark & \textbf{89.29} & \textbf{90.97} & \textbf{95.15} & \textbf{89.41} \\
\bottomrule
\end{tabular}
}
\label{tab:ablation}
\end{table}

To evaluate the contribution of ratio-based difficulty-aware (RDA) module and progressive adaptive loss weighting (PALW) mechanism in the proposed DPHR strategy, we conduct ablation study with MCCG as the baseline to analyze their impact on R@1 and AP metrics. The results on University-1652 are reported in Table~\ref{tab:ablation}.

From Table~\ref{tab:ablation}, it can be observed that the RDA module alone performs worse than the baseline, as emphasizing hard negatives too early amplifies noisy gradients from immature embeddings. In contrast, the PALW mechanism alone consistently improves performance by progressively suppressing early-stage noise and strengthening hard-negative mining as training stabilizes. When combined RDA and PALW, our DPHR achieves the best results, demonstrating their complementary roles. PALW ensures robust training in the early phase, while RDA enhances discriminability later, validating the necessity of their joint design.

\subsection{Visualization Analysis}

\begin{figure}[t]
    \centering
    \includegraphics[width=\linewidth]{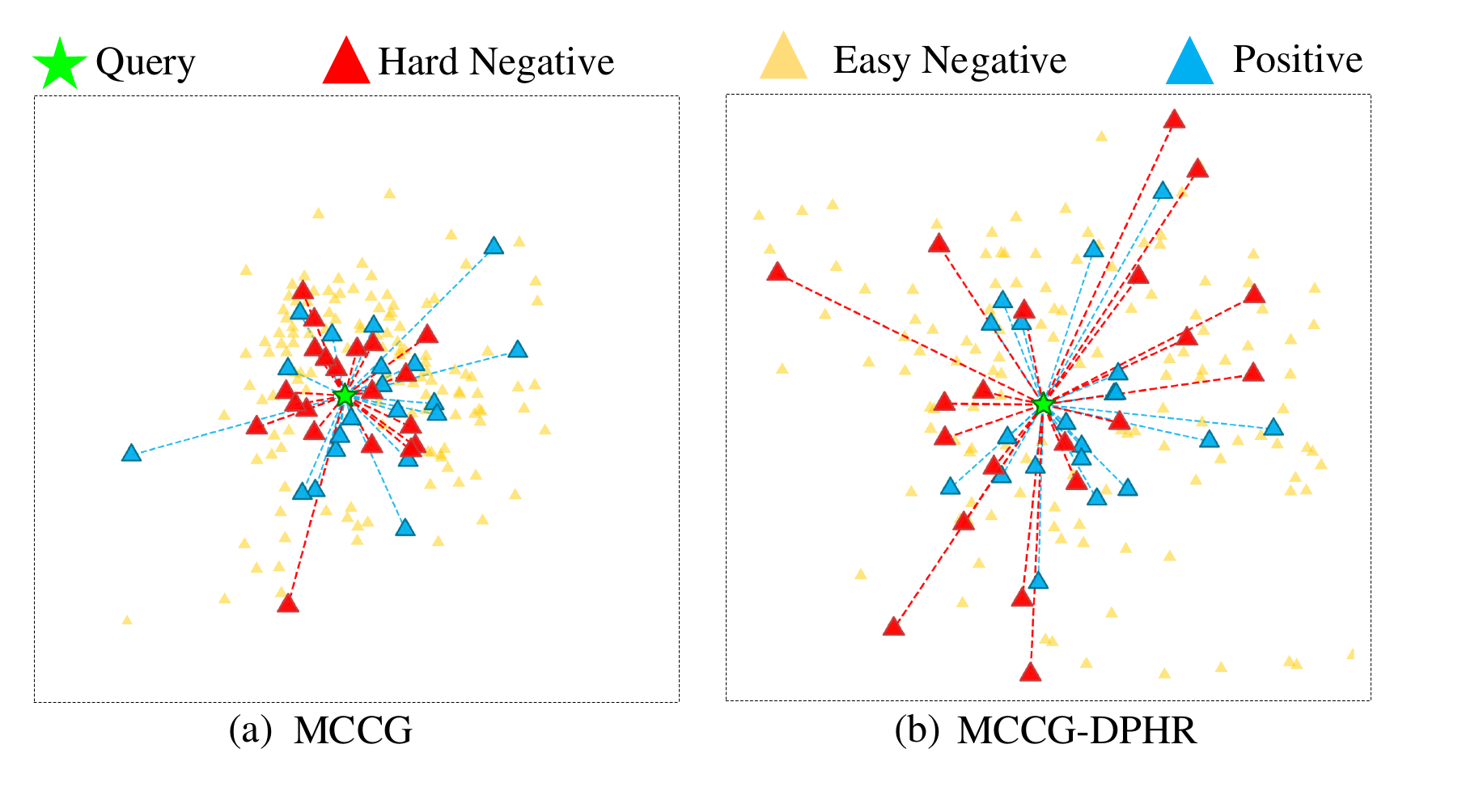}
    \caption{The t-SNE visualization comparing MCCG and MCCG-DPHR. Our strategy improves separation between queries and hard negatives, demonstrating its effectiveness in handling challenging negative samples.}
    \label{fig:tsne}
\end{figure}

To qualitatively assess the effectiveness of our proposed strategy in addressing hard negatives, we conducted a satellite-to-drone retrieval analysis using 20 randomly selected queries. For each query, the top-20 drone images were retrieved under both MCCG and MCCG-DPHR, and their distributions were visualized via t-SNE. We specifically highlight cases where MCCG incorrectly ranks a hard negative at Rank-1. To improve statistical reliability, the 20 individual plots were merged into a single visualization, with all queries fixed at the central position. As shown in Fig.~\ref{fig:tsne}, MCCG-DPHR pushes hard negatives farther from the queries, thereby yielding more accurate and robust retrieval results.

\section{Conclusion}

In this paper, we proposed a dual-level progressive hardness-aware reweighting strategy, namely DPHR, for CVGL task, which combines a sample-level ratio-based difficulty-aware module and a batch-level progressive adaptive loss weighting mechanism. The proposed method dynamically emphasizes hard negatives while suppressing early-stage noise. 
Extensive experiments on University-1652 and SUES-200 benchmarks demonstrate consistent performance improvement
across all retrieval directions and altitudes, which validates its effectiveness and robustness compared with state-of-the-art methods. 


\bibliographystyle{IEEEtran}
\bibliography{refs}

\begin{thebibliography}{10}
\providecommand{\url}[1]{#1}
\csname url@samestyle\endcsname
\providecommand{\newblock}{\relax}
\providecommand{\bibinfo}[2]{#2}
\providecommand{\BIBentrySTDinterwordspacing}{\spaceskip=0pt\relax}
\providecommand{\BIBentryALTinterwordstretchfactor}{4}
\providecommand{\BIBentryALTinterwordspacing}{\spaceskip=\fontdimen2\font plus
\BIBentryALTinterwordstretchfactor\fontdimen3\font minus \fontdimen4\font\relax}
\providecommand{\BIBforeignlanguage}[2]{{%
\expandafter\ifx\csname l@#1\endcsname\relax
\typeout{** WARNING: IEEEtran.bst: No hyphenation pattern has been}%
\typeout{** loaded for the language `#1'. Using the pattern for}%
\typeout{** the default language instead.}%
\else
\language=\csname l@#1\endcsname
\fi
#2}}
\providecommand{\BIBdecl}{\relax}
\BIBdecl

\bibitem{Lin2013cross}
T.-Y. Lin, S.~Belongie, and J.~Hays, ``{Cross-View Image Geolocalization},'' in \emph{Proceedings of the IEEE/CVF Conference on Computer Vision and Pattern Recognition (CVPR)}, 2013, pp. 891--898.

\bibitem{Zhu2018vision}
P.~Zhu, L.~Wen, X.~Bian, H.~Ling, and Q.~Hu, ``{Vision Meets Drones: A Challenge},'' \emph{arXiv preprint arXiv:1804.07437}, 2018.

\bibitem{Humenberger2022investigating}
M.~Humenberger, Y.~Cabon, N.~Pion \emph{et~al.}, ``{Investigating the Role of Image Retrieval for Visual Localization: An Exhaustive Benchmark},'' \emph{International Journal of Computer Vision}, vol. 130, no.~7, pp. 1811--1836, 2022.

\bibitem{yu2019building}
Q.~Yu, C.~Wang, B.~Cetiner \emph{et~al.}, ``Building {I}nformation {M}odeling and {C}lassification by {V}isual {L}earning {A}t {A} {C}ity {S}cale,'' \emph{arXiv preprint arXiv:1910.06391}, 2019.

\bibitem{Zheng2020University1652}
Z.~Zheng, Y.~Wei, and Y.~Yang, ``{University-1652}: A {M}ulti-view {M}ulti-source {B}enchmark for {D}rone-based {G}eo-localization,'' in \emph{Proceedings of the ACM International Conference on Multimedia (ACM MM)}, 2020, pp. 1395--1403.

\bibitem{triplet}
J.~Wang, Y.~Song, T.~Leung, C.~Rosenberg, J.~Wang, J.~Philbin, B.~Chen, and Y.~Wu, ``Learning fine-grained image similarity with deep ranking,'' in \emph{2014 IEEE Conference on Computer Vision and Pattern Recognition}, 2014, pp. 1386--1393.

\bibitem{soft-margin-triplet}
N.~N. Vo and J.~Hays, ``{Localizing and Orienting Street Views Using Overhead Imagery},'' in \emph{European conference on computer vision}.\hskip 1em plus 0.5em minus 0.4em\relax Springer, 2016, pp. 494--509.

\bibitem{infonce}
A.~v.~d. Oord, Y.~Li, and O.~Vinyals, ``Representation {L}earning with {C}ontrastive {P}redictive {C}oding,'' \emph{arXiv preprint arXiv:1807.03748}, 2018.

\bibitem{HADGEO}
C.~Li, C.~Yan, X.~Xiang \emph{et~al.}, ``{HADGEO}: Image-based {3-DoF} {C}ross-{V}iew {G}eo-{L}ocalization with {H}ard {S}ample {M}ining,'' in \emph{Proceedings of the IEEE International Conference on Acoustics, Speech and Signal Processing (ICASSP)}, 2024, pp. 3520--3524.

\bibitem{Sample4Geo}
F.~Deuser, K.~Habel, and N.~Oswald, ``{Sample4Geo}: {H}ard {N}egative {S}ampling for {C}ross-{V}iew {G}eo-{L}ocalisation,'' in \emph{Proceedings of the IEEE/CVF International Conference on Computer Vision (ICCV)}, 2023, pp. 16\,847--16\,856.

\bibitem{Effective-Negative-Sampling}
J.~Park, C.~Sung, S.~Lee, D.~Kang, and H.~Myung, ``{Cross-View Geo-Localization via Effective Negative Sampling},'' in \emph{Proceedings of International Conference on Control, Automation and Systems (ICCAS)}.\hskip 1em plus 0.5em minus 0.4em\relax IEEE, 2024, pp. 1078--1083.

\bibitem{Cai2019GroundToAerial}
S.~Cai, Y.~Guo, S.~Khan, J.~Hu, and G.~Wen, ``{Ground-to-Aerial Image Geo-Localization with a Hard Exemplar Reweighting Triplet Loss},'' in \emph{Proceedings of the IEEE/CVF International Conference on Computer Vision (ICCV)}, 2019, pp. 8391--8400.

\bibitem{Zhu2023SUES200}
R.~Zhu, L.~Yin, M.~Yang, F.~Wu, Y.~Yang, and W.~Hu, ``{SUES-200}: {A Multi-Height Multi-Scene Cross-View Image Benchmark Across Drone and Satellite},'' \emph{IEEE Transactions on Circuits and Systems for Video Technology}, vol.~33, no.~9, pp. 4825--4839, 2023.

\bibitem{Shen2023MCCG}
T.~Shen, Y.~Wei, L.~Kang, Wan \emph{et~al.}, ``{MCCG}: A {C}onvnext-{B}ased {M}ultiple-{C}lassifier {M}ethod for {C}ross-{V}iew {G}eo-{L}ocalization,'' \emph{IEEE Transactions on Circuits and Systems for Video Technology}, vol.~34, no.~3, pp. 1456--1468, 2023.

\bibitem{ConvNeXt-based}
F.~Guan, N.~Zhao, Z.~Fang, L.~Jiang, J.~Zhang, Y.~Yu, and H.~Huang, ``{Multi-level Representation Learning Via ConvNeXt-based Network for Unaligned Cross-View Matching},'' \emph{Geo-spatial Information Science}, pp. 1--14, 2025.

\bibitem{ConvNeXt-shaa}
N.~Chen, D.~Zhang, K.~Jiang, M.~Yu, Y.~Zhu, T.-s. Lou, and L.~Zhao, ``{SHAA: Spatial Hybrid Attention Network with Adaptive Cross-Entropy Loss Function for UAV-view Geo-localization},'' \emph{IEEE Transactions on Circuits and Systems for Video Technology}, 2025.

\bibitem{LPN}
T.~Wang, Z.~Zheng, C.~Yan, J.~Zhang, Y.~Sun, B.~Zheng, and Y.~Yang, ``{Each part matters: Local Patterns Facilitate Cross-View Geo-Localization},'' \emph{IEEE Transactions on Circuits and Systems for Video Technology}, vol.~32, no.~2, pp. 867--879, 2021.

\bibitem{University-method02}
W.~Gan, Y.~Zhou, X.~Hu \emph{et~al.}, ``Learning {R}obust {F}eature {R}epresentation for {C}ross-{V}iew {I}mage {G}eo-{L}ocalization,'' \emph{IEEE Geoscience and Remote Sensing Letters}, 2025.

\end{thebibliography}

\end{document}